\documentclass[journal]{IEEEtranTIE}
\usepackage[pdftex]{graphicx}
\usepackage{cite}
\usepackage{picinpar}
\usepackage{amsmath}
\usepackage{url}
\usepackage{flushend}
\usepackage{colortbl}
\usepackage{soul}
\usepackage{multirow}
\usepackage{pifont}
\usepackage{color}
\usepackage{alltt}
\usepackage[hidelinks]{hyperref}
\usepackage{enumerate}
\usepackage{siunitx}
\usepackage{breakurl}
\usepackage{epstopdf}
\usepackage{pbox}
\usepackage{tcolorbox}
\usepackage{textcomp, gensymb}
\usepackage{amsfonts}
\usepackage{stfloats}
\usepackage{algorithm2e}
\usepackage[siunitx, RPvoltages]{circuitikz}
\usepackage{tikz}
\usepackage[inline]{enumitem}
\usepackage{booktabs}
\usepackage[utf8]{inputenc}
\usepackage{xcolor}
\usepackage{balance}
\usepackage{tabularray}
\usepackage{colortbl}
\usetikzlibrary{shapes.geometric}
\usetikzlibrary{positioning}
\DeclareMathOperator*{\argmin}{arg\,min}

\begin{document}
\title{A Self-Commissioning Edge Computing Method for Data-Driven Anomaly Detection in Power Electronic Systems}

\author{
	\vskip 1em

	Pere~Izquierdo~G\'{o}mez, \emph{Student Member,~IEEE},
	Miguel~E.~L\'{o}pez Gajardo, \emph{Student Member,~IEEE},
	Nenad~Mijatovic, \emph{Senior Member,~IEEE},
	and Tomislav~Dragi\v{c}evi\'{c}, \emph{Senior Member,~IEEE}

	\thanks{

	The authors are with the Department of Wind and Energy Systems, Technical University of Denmark, Kongens Lyngby 2800, Denmark (email: \{pizgo, melga, nemi, tomdr\}@dtu.dk).

	}
}

\maketitle

\begin{abstract}
	Ensuring the reliability of power electronic converters is a matter of great importance, and data-driven condition monitoring techniques are cementing themselves as an important tool for this purpose.
	However, translating methods that work well in controlled lab environments to field applications presents significant challenges, notably because of the limited diversity and accuracy of the lab training data.
	By enabling the use of field data, online machine learning can be a powerful tool to overcome this problem, but it introduces additional challenges in ensuring the stability and predictability of the training processes.
	This work presents an edge computing method that mitigates these shortcomings with minimal additional memory usage, by employing an autonomous algorithm that prioritizes the storage of training samples with larger prediction errors.
	The method is demonstrated on the use case of a self-commissioning condition monitoring system, in the form of a thermal anomaly detection scheme for a variable frequency motor drive, where the algorithm self-learned to distinguish normal and anomalous operation with minimal prior knowledge.
	The obtained results, based on experimental data, show a significant improvement in prediction accuracy and training speed, when compared to equivalent models trained online without the proposed data selection process.
\end{abstract}


\markboth{IEEE TRANSACTIONS ON INDUSTRIAL ELECTRONICS}%
{}

\definecolor{limegreen}{rgb}{0.2, 0.8, 0.2}
\definecolor{forestgreen}{rgb}{0.13, 0.55, 0.13}
\definecolor{greenhtml}{rgb}{0.0, 0.5, 0.0}

\section{Introduction}
\IEEEPARstart{P}{ower} electronic converters play a key role in the electrification of the global energy sector and are therefore becoming ubiquitous in a wide range of applications.
It is thus fundamental to ensure that power converters can cost-effectively satisfy reliability requirements over their designed operating lifetime, especially in safety-critical areas such as transportation, water supply, industrial, and power transmission systems.
In less critical environments, maximizing reliability can also result in significant improvements, by reducing system downtime and its associated costs.

Increasing the reliability of power converters in the design phase often involves oversizing their components or introducing redundancies; but these modifications may not always be economically viable~\cite{Avenas2015}.
An alternative approach to failure minimization is to employ condition monitoring techniques, which can inform the scheduling of predictive maintenance by assessing the health status of the systems under consideration.
Condition monitoring analyzes the changes and trends of a set of device characteristics in order to identify failure precursors and thus anticipate the need for maintenance before serious deterioration or breakdown occurs~\cite{Han2003}.
Successful implementations of condition monitoring methods for power electronic converters must overcome significant challenges, which can be broadly summarized as follows:
\begin{enumerate}[wide]
    \item Condition monitoring methods are typically limited in scope to a single type of component. However, at the application level, the interactions between the multiple components of a converter often invalidate the assumption that the degradation of each of them can be analyzed independently~\cite{Wang2021}.
    \item In practical applications, it is often challenging to directly measure device parameters indicative of degradation. For example, aging failures in power semiconductor devices typically occur due to wire-bonding contact degradation and the creeping of the baseplate solder layer~\cite{Iannuzzo2014}. Mechanical stress gauges and localized electrical resistance measurements within a device or module can directly monitor these mechanisms. Nonetheless, these additional sensors increase costs and are challenging to install and maintain in commercial products, which makes indicators derived from device or converter-level measurements more attractive in practical use cases~\cite{Yang2010}.
    \item The degradation dynamics observed under controlled test conditions often cannot be extrapolated to field operation, where power converter systems experience substantially more diverse stress profiles~\cite{Sangwongwanich2020}. The broad variety of real-world influences, which may range from control techniques to environmental conditions (e.g.\ ambient temperature, humidity), virtually forbids exhaustive testing within the entirety of the system's operating space.
\end{enumerate}

There is therefore a need for condition monitoring solutions that operate at the converter level, require few or no additional sensors, and are able to achieve precise prognostics and health management performance under a wide variety of operating conditions.
The present article proposes one such method, employing a self-commissioning data-driven model trained in an online manner to detect thermal anomalies in a power converter, and able to run in its entirety with minimal prior knowledge and on an edge device.

\subsection{Failure and Degradation in Power Electronic Systems}
As a result of the aforementioned limitations, application-level monitoring techniques often focus on a few dominant failure and degradation mechanisms~\cite{Wang2021}.
Accordingly, it is fundamental to identify these dominant mechanisms and to construct relevant metrics.

In a 2011 industry-based survey of reliability in power electronic converters~\cite{Yang2011}, respondents ranked power devices as the most fragile components, followed by capacitors, gate drives, and connectors.
Failures associated with inductors and resistors were comparatively rare. The~\textit{Handbook of electronic package design}~\cite{pecht1991a} instead identifies capacitors as the component most prone to failure, followed by printed circuit boards (PCBs), semiconductors, soldering, and connectors.

Field experiences in photovoltaic (PV) systems analyzed in~\cite{Hacke2018} indicated that inverters are the first cause of service tickets in plants requiring maintenance, accounting for up to 70\% of requests.
At the component level, software errors consistently ranked first.
However, conclusions were not as clear regarding hardware malfunctions: cards and boards, cooling systems and fans, ac contactors and circuit breakers, and insulated-gate bipolar transistors (IGBTs) all ranked highly, but none of them stood out consistently.
Another study on PV systems~\cite{Golnas2013} reached similar conclusions, with cards and boards accounting for 13\% of hardware-related issues, followed by ac contactors with 12\%, fans with 6\%, and matrices and IGBTs with 6\%.
A~2019 study on the reliability of power converters in wind turbines~\cite{Fischer2019} found phase modules to form the largest portion of failed converter components, with a 22\% of the total.
These were followed by control boards, cooling systems, and main circuit breakers.

Despite humidity-related degradation mechanisms receiving increased attention in recent years~\cite{Fischer2019, Sadik2017, Kremp2018}, consensus in the literature points to thermal stresses as the most dominant factor in the reliability of power modules~\cite{Wu1995, Choi2015, Durand2016, Ciappa2002}.
Bond wire lifting is regarded as the main failure mechanism in IGBT modules, followed by solder joint fatigue.
Both mechanisms have been found to be directly tied to thermal cycling.
\cite{pecht1991a} identifies ``Temperature --- Steady State \& Cyclical'' as the main stress factor in 55\% of recorded failures, ahead of ``Vibration/Shock'' (20\%), ``Humidity/Moisture'' (19\%), and ``Contaminants \& Dust'' (6\%).
Temperature swings and steady-state temperatures both play a crucial role in component degradation, affecting bond wire lifting and solder lifetime~\cite{Bouarroudj2008, Bayerer2008, Choi2017, Zeng2019}.

Classical condition indicators of power modules are commonly based on forward voltage and junction-to-case thermal impedance~\cite{Perpinya2012, Cova1998, Dupont2007}.
A deviation in on-state voltage or junction-to-case thermal resistance reflects in an increase in junction temperature of a few \textcelsius~\cite{Smet2011}.
A condition monitoring technique may exploit this relationship to extract information on the status of the device; however, changes in junction temperature will typically reflect in even smaller change in case temperature~\cite{Xiang2011}, which may be challenging to measure accurately.
Nevertheless, mean junction temperature is an important factor in determining the lifetime of power modules.
According to~\cite{Zeng2019}, with a power cycling test at \qty{50}{Hz} and junction temperature swings between \qty{23}{\celsius} and \qty{27}{\celsius}, a \qty{10}{\celsius}~reduction in mean junction temperature could result in a tenfold increase in the lifetime of the IGBT modules under test.

Most other components in power electronic converters, including capacitors, integrated circuits, PCBs, and connectors, also exhibit an inversely proportional relationship between their average temperature and their expected lifetime~\cite{Huai2014}.
The cooling system of power converters, whether active or passive, is responsible for maintaining the average temperature of the components within its designed boundaries.
Therefore, ensuring that it remains in adequate condition throughout the lifetime of the converter can result in significant improvements in the reliability of its components.

\subsection{Anomaly Detection}
Anomaly detection refers to the process of identifying patterns in data that do not conform to expected behavior~\cite{Prasad2009}.
Most anomaly detection methods aim to first characterize the healthy operation of a system, and then to identify instances that do not conform to it.
This can be achieved using model-free or model-based approaches, both of which have found numerous applications in power electronics~\cite{Kang2018}.

Model-free anomaly detection typically employs data clustering methods to directly quantify the difference between anomalous samples and a given healthy data distribution.
Some examples within the field of power electronics include~\cite{Yu2022}, where the authors employ data-driven graph representations to detect anomalies in correlated sensors; and~\cite{Jin2012}, which presents a method for health monitoring of cooling fans using high-dimensional feature extraction and a Mahalanobis distance criterion.
However, model-free methods may fail to sufficiently exploit known relevant relationships in the data, e.g.\ the correlation between a device's current and its power losses.

Alternatively, model-based anomaly detection commonly relies on a model of the monitored system under healthy conditions, and seeks to identify anomalous behavior by comparing system measurements with model predictions, thereby analyzing prediction errors or residuals.
The underlying assumption in model-based methods is that, for a given set of input data, the model will make predictions corresponding to the expected operation.
The predictions will then match observations more closely for healthy operation than for degraded operation, allowing for the identification of unexpected behavior when prediction residuals increase.
Model-based condition monitoring approaches are particularly suited to converter systems under variable load and ambient conditions~\cite{Yang2010}, as models can be trained to make predictions only on relevant input-output relationships and thus achieve increased robustness.
Some examples of model-based anomaly detection in power electronics include~\cite{Baker2022}, where a recurrent neural network is trained to identify failures in multi-level inverters; and~\cite{Jiang2020}, in which Gaussian process regression is employed with a genetic algorithm to identify the status of a dc-to-dc converter.

% --------------------------------------------------------------------------------
\section{System Description}\label{sec:system_description}

\begin{figure}
    \centering
    \ctikzset{sources/scale=0.7}
\ctikzset{diodes/scale=0.4}
\ctikzset{inductors/scale=0.5}
\ctikzset{capacitors/scale=0.5}
\ctikzset{transistors/scale=0.5}
\ctikzset{bipoles/border margin=1} 
\ctikzset{nodes width/.initial=.02}
\ctikzset{current arrow scale=24}

\begin{circuitikz}[american]
%\draw[step=0.5cm,gray,very thin] (0,0) grid (10,6);
% Rectifier side
    % Voltage source
    \draw (0, 4) node[vsourcesinshape, rotate=90, label={[font=\footnotesize]east:Grid}] (Vgrid){};
    % Upper diodes
    \draw (0.25, 4.15) -- ++(0.35, 0.35) to [short, -*] (1, 4.5) to [diode] ++(0, 1) -- (2, 5.5);
    \draw (Vgrid.270) to [short, -*] (1.5, 4) -- ++(0, 0.5) to [diode, -*] ++(0, 1);
    \draw (0.25, 3.85) -- ++(0.35, -0.35) to [short, -*] (2, 3.5) -- ++(0, 1) to [diode, -*] ++(0, 1);
    % Lower diodes
    \draw (2, 2.5) to [short] (1, 2.5) to [diode] (1, 3.5) -- (1, 4.5) ++(0, 1);
    \draw [short, *-] (1.5, 2.5) to [diode] (1.5, 3.5) -- (1.5, 4.5);
    \draw [short, *-] (2, 2.5) to [diode] (2, 3.5) -- (2, 4.5);

% DC link
    \draw (2, 5.5) to [inductor, -*, name=topL] (3.25, 5.5);
    \draw (2.625, 5.55) node[label={\footnotesize $L_{dc}$}] {};
    \draw[thick] (topL.core west) -- (topL.core east);
    \draw (2, 2.5) to [inductor, -*, name=bottomL] (3.25, 2.5); 
    \draw[thick] (bottomL.core west) -- (bottomL.core east);
    \draw (3.25, 5.5) to [capacitor, l_={\footnotesize $C_{dc}$}] (3.25, 2.5);
    
% Inverter side
    % First leg
    \draw (3.25, 5.5) to [short, -*] (4, 5.5);
    \draw (4, 5.5) -- (4, 5.375) node[nigbt, bodydiode, anchor=C](igbt1){};
    \draw (igbt1.E) to [short, -*] (4, 4.5) -- (4, 3.375) node[nigbt, bodydiode, anchor=C](igbt2){};
    \draw (igbt2.E) to [short, -*] (4, 2.5);
    \draw (3.25, 2.5) -- (4, 2.5);
    
    % Second leg
    \draw (4, 5.5) to [short, -*] (4.75, 5.5);
    \draw (4.75, 5.5) -- (4.75, 5.375) node[nigbt, bodydiode, anchor=C](igbt3){};
    \draw (igbt3.E) to [short, -*] (4.75, 4) -- (4.75, 3.375) node[nigbt, bodydiode, anchor=C](igbt4){};
    \draw (igbt4.E) to [short, -*] (4.75, 2.5);
    \draw (4, 2.5) -- (4.75, 2.5);
    
    % Third leg
    \draw (4, 5.5) to [short, -*] (4.75, 5.5);
    \draw (4.75, 5.5) -- (5.5, 5.5) -- (5.5, 5.375) node[nigbt, bodydiode, anchor=C](igbt5){};
    \draw (igbt5.E) to [short, -*] (5.5, 3.5) -- (5.5, 3.375) node[nigbt, bodydiode, anchor=C](igbt6){};
    \draw (4.75, 2.5) -- (5.5, 2.5) -- (igbt6.E);
    
    % Motor
    \draw (6.5, 4) node[elmech, scale=0.7](motor){M\textsubscript{1}};
    \draw (4, 4.5) -- (6, 4.5) -- (motor.135);
    \draw (4.75, 4) -- (motor.left);
    \draw (5.5, 3.5) -- (6, 3.5) -- (motor.225);
    
    % Motor coupling
    \draw (7.5, 4) node[elmech, scale=0.7](motor_load){M\textsubscript{2}};
    \draw (motor.7.5) -- (motor_load.172.5);
    \draw (motor.-7.5) -- (motor_load.187.5);
    \draw[dashed] (motor_load.right) -- ++(0.35, 0) -- ++(0.15, 0);
    \draw[dashed] (motor_load.30) -- ++(0.35, 0.35) -- ++(0.15, 0);
    \draw[dashed] (motor_load.-30) -- ++(0.35, -0.35) -- ++(0.15, 0);
    
\end{circuitikz}
    \vspace{-0.7cm}
    \caption{Simplified circuit diagram of the monitored frequency converter in the experimental setup, as part of a Danfoss FC 302 variable-frequency motor drive.}\label{fig:lab_setup_circuit_diagram}
\end{figure}
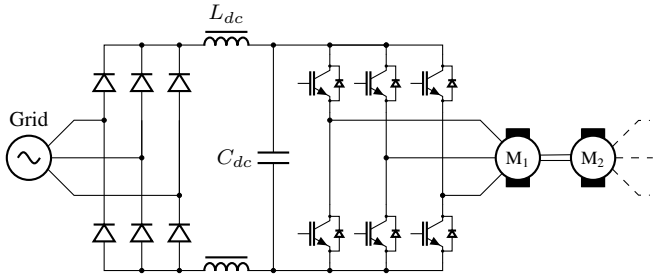

\begin{figure}
    \centering
    \includegraphics[width=\columnwidth]{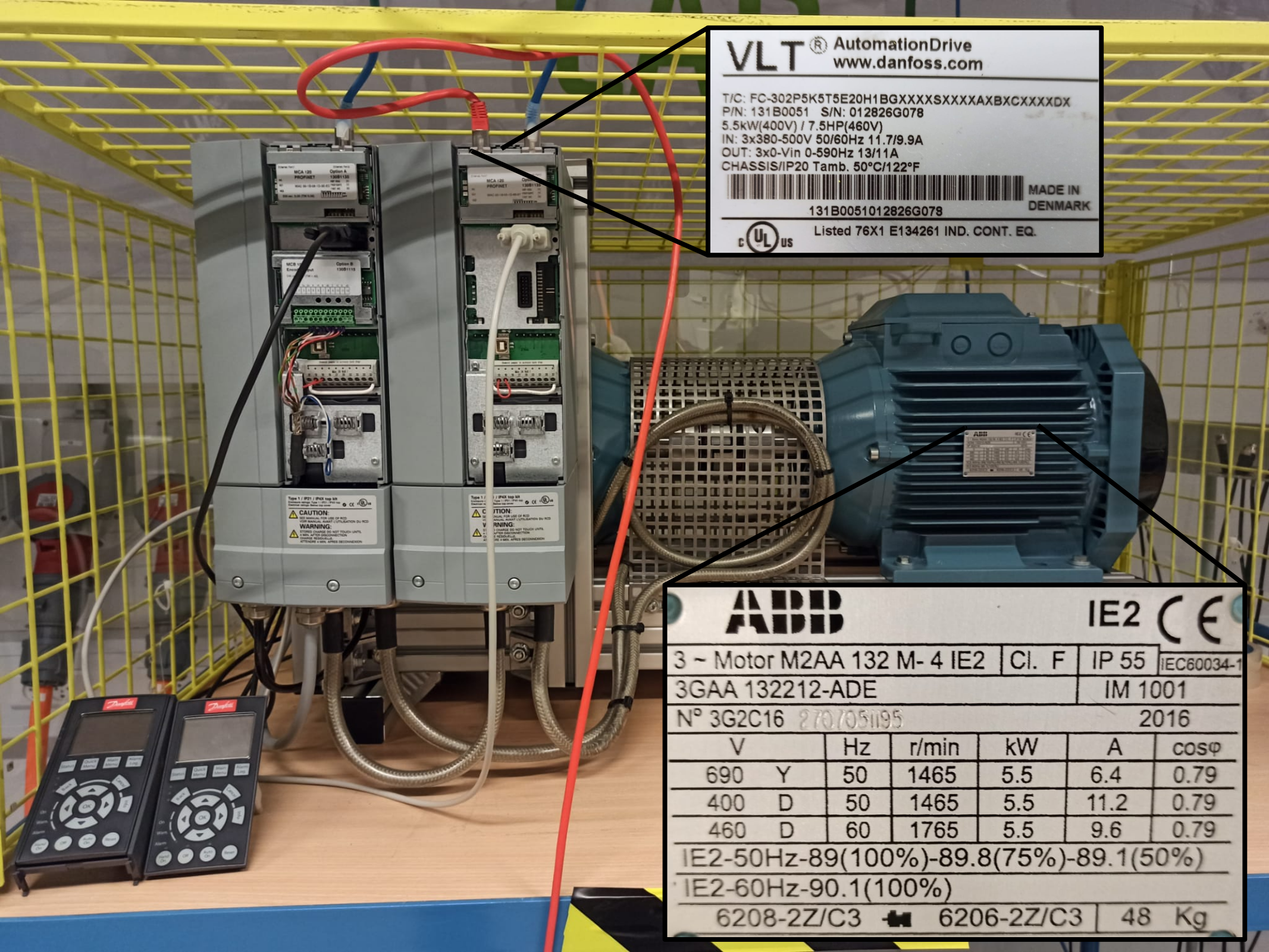}
    \caption{Picture of the motor test bench used to collect experimental data.}\label{fig:lab_setup_photo}
\end{figure}

The proposed methodology is applied to the detection of thermal anomalies in a variable-frequency power converter, subjected to a variety of operating conditions.
A data set is collected from a low-voltage motor drive test bench, which consists of two \qty{5.5}{kW} ABB 3-phase induction machines coupled to the same shaft, and driven by two low-voltage Danfoss VLT\textsuperscript{\tiny{\textregistered}} AutomationDrive FC302 variable-frequency converters, also rated at \qty{5.5}{kW}.
A simplified circuit diagram of the power electronic converter is shown in Fig.~\ref{fig:lab_setup_circuit_diagram}, and a photograph of the experimental test bench is shown in Fig.~\ref{fig:lab_setup_photo}.

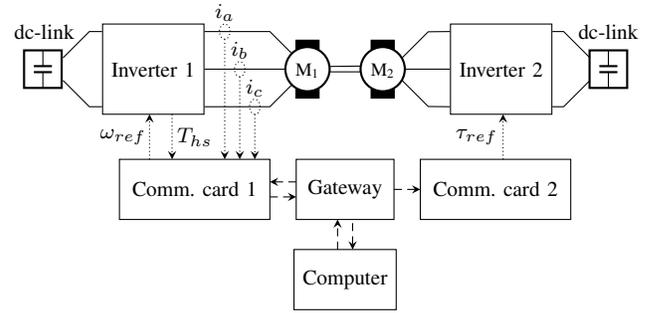
\begin{figure}
    \centering
    \ctikzset{sources/scale=0.7}

\newcommand\dclink{}
\def\dclink[#1](#2,#3){  % [draw options](center)
    \draw[#1] (#2, #3) ++(-0, 0.25) to [/tikz/circuitikz/bipoles/length=0.4cm, capacitor] ++(0, -0.5);
    \draw[#1, thick] (#2, #3) ++(-0.25, -0.25) -- ++(0, 0.5) -- ++(0.5, 0) -- ++(0, -0.5) -- cycle;
}

\begin{circuitikz}[american]
    % Left-side dc-link
    \dclink[](0, 4) {\textsubscript{dc-link}};
    \node[align=center] at (0, 4.5) {\footnotesize dc-link};
    
    % Left inverter and motor
    \node [draw, minimum width=1.2cm, minimum height=1.2cm] (drive1) at (1.45, 4) {\footnotesize Inverter 1};
    \draw (0.25, 4.15) -- ++(0.35, 0.35) -- ([yshift=0.5cm]drive1.180);
    \draw (0.25, 3.85) -- ++(0.35, -0.35) -- ([yshift=-0.5cm]drive1.180);

    \draw (3.5, 4) node[elmech, scale=0.7](motor1){M\textsubscript{1}};
    \draw ([yshift=0.5cm]drive1.0) -- (3, 4.5) -- (motor1.135);
    \draw (drive1.0) -- (3, 4) -- (motor1.left);
    \draw ([yshift=-0.5cm]drive1.0) -- (3, 3.5) -- (motor1.225);
    
    % Right motor and coupling
    \draw (4.5, 4) node[elmech, scale=0.7](motor2){M\textsubscript{2}};
    \draw (motor1.7.5) -- (motor2.172.5);
    \draw (motor1.-7.5) -- (motor2.187.5);
    
    % Right drive
    \node [draw, minimum width=1.2cm, minimum height=1.2cm, right=0.6cm of motor2] (drive2) {\footnotesize Inverter 2};
    \draw (motor2.right) -- ++(0.35, 0) -- (drive2.180);
    \draw (motor2.30) -- ++(0.35, 0.35) -- ([yshift=0.5cm]drive2.180);
    \draw (motor2.-30) -- ++(0.35, -0.35) -- ([yshift=-0.5cm]drive2.180);
    
    % Right dc-link
    \dclink[](7.5, 4);
    \node[align=center] at (7.5, 4.5) {\footnotesize dc-link};
    \draw ([yshift=0.5cm]drive2.0) -- ++(0.15, 0) -- ++(0.35, -0.35);
    \draw ([yshift=-0.5cm]drive2.0) -- ++(0.15, 0) -- ++(0.35, 0.35);
    
    % Communication card
    \node [draw, align=center,  minimum width=2cm, minimum height=0.8cm] (comm_card_1) at (2, 2.4) {\footnotesize Comm. card 1};
    
    % Current sensors
    \draw[densely dotted] (2.4, 4.5) ellipse (0.07 and 0.1);
    \draw[densely dotted, -stealth] (2.4, 4.4) -- (2.4, 2.8);
    \node[] at (2.4, 4.75) {\footnotesize $i_{a}$};
    \draw[densely dotted] (2.6, 4) ellipse (0.07 and 0.1);
    \draw[densely dotted, -stealth] (2.6, 3.9) -- (2.6, 2.8);
    \node[] at (2.6, 4.25) {\footnotesize $i_b$};
    \draw[densely dotted] (2.8, 3.5) ellipse (0.07 and 0.1);
    \draw[densely dotted, -stealth] (2.8, 3.4) -- (2.8, 2.8);
    \node[] at (2.8, 3.75) {\footnotesize $i_c$};
    
    % Other communications (drive 1)
    \draw[densely dotted, -stealth] (1.7, 3.4) -- (1.7, 2.8);
    \node[] at (2, 3.1) {\footnotesize $T_{hs}$};
    \draw[densely dotted, -stealth] (1.4, 2.8) -- (1.4, 3.4);
    \node[] at (1.05, 3.1) {\footnotesize $\omega_{ref}$};
    
    % Gateway
    \node[draw, align=center,  minimum width=1.3cm, minimum height=0.8cm] (gateway) at (4, 2.4) {\footnotesize Gateway};
    \draw[densely dashed, stealth-] (3, 2.5) -- (3.35, 2.5);
    \draw[densely dashed, -stealth] (3, 2.3) -- (3.35, 2.3);
    
    % Communication card 2
    \node [draw, align=center,  minimum width=2cm, minimum height=0.8cm] (comm_card_2) at (6, 2.4) {\footnotesize Comm. card 2};
    \draw[densely dashed, -stealth] (4.65, 2.4) -- (5, 2.4);
    \draw[densely dotted, -stealth] (6.1, 2.8) -- (6.1, 3.4);
    \node[] at (5.75, 3.1) {\footnotesize $\tau_{ref}$};
    
    % Computer
    \node [draw, align=center,  minimum width=1cm, minimum height=0.8cm] (computer) at (4, 1.2) {\footnotesize Computer};
    \draw[densely dashed, -stealth] (3.9, 1.6) -- (3.9, 2);
    \draw[densely dashed, stealth-] (4.1, 1.6) -- (4.1, 2);
\end{circuitikz}
    \vspace{-0.7cm}
    \caption{Block diagram depicting the experimental test bench. Solid lines represent electrical connections, dotted lines represent drive-internal measurements and communication, and dashed lines represent external communication. Motor 1 (M\textsubscript{1}) is speed-controlled, while motor 2 (M\textsubscript{2}) is torque-controlled. The two inverters are connected to a common dc-link.}\label{fig:lab_setup_block_diagram}
\end{figure}

A communication card is integrated to each of the two drives, allowing for data transmission via Profinet.
Information is sent to and from a network gateway, which in turn allows for communication with a computer.
For the anomaly detection process, the logged variables from the drive include the phase rms currents $i_a$, $i_b$, $i_c$, and the heat sink temperature $T_{hs}$.
The motor connected to the monitored converter is speed-controlled, while the second motor is torque-controlled, acting as a controllable load.
A block diagram representing the experimental setup is shown in Fig.~\ref{fig:lab_setup_block_diagram}.

The drive reference values for speed $\omega_{ref}$ and torque $\tau_{ref}$ are generated by random sampling from the operating range of the motor setup, as
\begin{equation}
    \omega_{ref} \sim U(0, 1465) \text{ [r/min]}, \quad \tau_{ref} \sim U(0, 28) \text{ [N{$\cdot$}m]},
\end{equation}
where $U$ stands for the uniform distribution.
After sending the two references, they are held constant for a total of 10 minutes, after which there is a 50\% chance that a new randomized reference is sent, and a 50\% chance that the drives are stopped, allowing for the converters to cool down.
This process ensures that the collected data includes diverse heat sink temperature profiles, resulting from variable loading conditions.
The measurements are downsampled to a constant sampling period of 10 seconds, without interpolation.
The experimental data set was collected prior to the training process, allowing for algorithm benchmarking, cross-validation analysis, and facilitating parameter tuning.
The collected data set is then streamed to train the selected models in an online manner.

\begin{figure}
    \centering
    \includegraphics[scale=1]{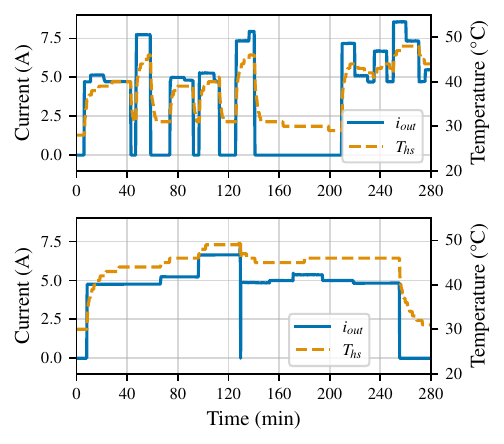}
    \vspace{-0.0cm}
    \caption{A portion of the collected data set, for \textit{normal} operation (top) and \textit{anomalous} operation (bottom). The output RMS average phase current $i_{out}$ (blue line, solid) is plotted on the left axes, while the heat sink temperature $T_{hs}$ (orange line, dashed) is plotted on the right axes.}\label{fig:example_measurements}
\end{figure}

In order to gather data corresponding to thermal anomalies in the motor drive, the collection process included periods in which the air outlet of its cooling system was partially blocked.
Besides failure in the cooling system, other scenarios such as loss of phase and large asymmetries may also result in unexpected thermal profiles.
The complete recorded data set contains approximately 26 hours of data recorded during \textit{normal} operation, and 21 hours of data recorded during \textit{anomalous} operation.
The data corresponding to normal operation are employed in the training data set, while anomalous data are used only for validation.
For illustrative purposes, a graph displaying a portion of the collected data is shown in Fig.~\ref{fig:example_measurements}.

It is worth noting that the heat sink temperature measurements obtained from the built-in sensor have a resolution of only \qty{1}{\celsius}.
This would hinder their representation with linear and autoregressive (AR) models, as consecutive measurements often result in the same temperature value, further motivating the use of data-driven nonlinear models.

% --------------------------------------------------------------------------------
\section{Proposed Methodology}\label{sec:methodology}
The proposed methodology is based on the online training of a neural network model using streaming data from a motor drive system, to then employ its predictions for anomaly detection.
With few documented exceptions (e.g.~\cite{Rafferty2018, Alam2020}), anomaly detection algorithms for electrical systems typically rely on the collection of static data sets, with model-based approaches employing these data for offline training.
In this manner, the designer has the opportunity to verify that the trained models produce accurate predictions on the complete data set.
However, it can be time-consuming and costly to collect sufficiently representative data in a controlled environment, as the conditions present in the controlled environment may not accurately mirror those found in the field.

\subsection{Online Machine Learning}
Alternatively, data-driven models may be trained on real-time streaming data, in a process referred to as \emph{online} or \emph{incremental} machine learning.
Data from field operation can be collected in real-time and used to update the model parameters, accurately reflecting the behavior of the system under its current operating conditions.
This process can take place on a cloud platform or locally, where models are updated directly on the edge devices.
Cloud platforms enable the aggregation of large data sets, increasing the amount of data available at each training iteration.
However, cloud computation, data storage, and data transmission can carry significant economic costs, mostly involving charges from the cloud service provider and from the telecommunications system provider that maintains a reliable connection between each edge device and the internet.
On the other hand, model training on the edge devices carries virtually no additional costs, besides the upfront price of the local edge device hardware.
Additionally, it may even entirely remove the need for a connection to the internet, thereby further reducing costs and increasing the reliability of the monitoring system.

Despite these advantages, edge devices are much more limited than cloud platforms in terms of memory and computational resources.
In practice, this often means that the training process must take place in a continual setting, where only a small subset of the observed measurements can be held in memory at any given time, which often leads to instability in the online training process.
For instance, neural network models tend to experience \emph{catastrophic forgetting}, where models continuously overfit to their most recent training data and therefore lose their ability to generalize to older observations~\cite{Parisi2019}.
In the context of online regression, a neural network model is trained on a fixed-size data buffer $\mathcal{B}$, which without a sample selection mechanism contains the $|\mathcal{B}|$ most recent input-output pairs.
Training until convergence on $\mathcal{B}$ with a mean squared error loss will result in a model that generates approximately optimal predictions in a least squares sense.
However, this desirable property only holds for the current time buffer $\mathcal{B}_t$, and since the model has a limited capacity, minimizing the loss in $\mathcal{B}_t$ will increase the loss on the previous buffers $\mathcal{B}_0, \dots, \mathcal{B}_{t-1}$, i.e., it will lead to catastrophic forgetting.
Stopping the training on each buffer before convergence, e.g. by decreasing the learning rate(s), reducing the number of training epochs per buffer, or increasing the size of the model, can alleviate this problem; but it introduces a trade-off between the stability and the plasticity of the model, and can therefore lead to underfitting.
Numerous techniques have been developed to mitigate the effects of catastrophic forgetting in online learning settings while addressing the stability-plasticity problem, including:
\begin{enumerate*}[label=(\roman*), itemjoin={{; }}, itemjoin*={{; and }}]
    \item memory replay~\cite{Rebuffi2016}
    \item regularization~\cite{Kirkpatrick2017,Schwarz2018,He2021}
    \item knowledge distillation~\cite{Li2018}
    \item ensemble methods~\cite{Polikar2001,Doan2022}
    \item flexible model architectures~\cite{Draelos2016,Li2021}.
\end{enumerate*}
The interested reader may refer to~\cite{DeLange2022} for a comprehensive review of continual learning methods developed up to 2021.

Much of the existing literature focuses on the study of classification problems in task-incremental settings, assumes knowledge on when tasks change, and/or employs additional or growing models, either to learn new tasks or to generate artificial samples with deep generative architectures.
Many of these approaches are therefore not directly applicable in computationally resource-constrained industrial environments, and for regression tasks with minimal previous knowledge.

This article proposes a new algorithm for improved online training by mitigating catastrophic forgetting, with minimal additional computation and memory usage.
The method is based on the following idea: given a buffer of training data in an online regression setting, the samples with a lower prediction error contain less relevant information than those with a higher error, as the model is already able to generate accurate estimates given the corresponding inputs.
During online training, each new observation is set to replace the element in the buffer with the lowest prediction error, rather than the oldest element.
This method is closely related to \emph{prioritized experience replay}~\cite{Schaul2015}, commonly used in the context of deep reinforcement learning.
In prioritized experience replay, errors are stored in the data buffer, and data points with larger associated errors are sampled more frequently during training steps.
Although prioritized experience replay has been shown to outperform uniform sampling in terms of generalization and training speed, it is commonly assumed that the maximum buffer size is large enough to cover the entire observation space, and there is therefore little or no need to overwrite existing data.
The results presented in Section~\ref{sec:results} suggest that similar improvements can be obtained with the proposed method, where the buffer size is constant and limited.
The sample selection method in the context of online neural network training is shown in Algorithm~\ref{alg:selection}.

\RestyleAlgo{ruled}
\begin{algorithm}
\caption{Online training of a neural network with buffer sample selection.}\label{alg:selection}
\textit{\textbf{Initialization:}}\\
Initialize a neural network model $f \colon \mathbb{R}^{n_f} \rightarrow \mathbb{R}^{n_o}$ mapping $n_f$ sample features to $n_o$ outputs, containing $n_l$ layers with weights $\mathbf{W}^{(l)}$ and biases $\mathbf{b}^{(l)}$, $\forall l \in \{1, \ldots, n_l\}$. \\
Allocate a data buffer $\mathcal{B} = {\{\left(\mathbf{x}_i, \mathbf{y}_i\right)\}}_{i=1}^N$. \\
Define a training loss function $\ell \colon \mathbb{R}^{n_o} \times \mathbb{R}^{n_o} \rightarrow \mathbb{R}$. \\
Define a training time period limit $t_{\max}$. \\
\While{$t < t_{\max}$} {
\textit{\textbf{Buffer update:}}\\
Receive a new data pair $\mathbf{x}_{in} \in \mathbb{R}^{n_f}$, $\mathbf{y}_{in} \in \mathbb{R}^{n_o}$. \\
\eIf{$t < N$} {
    Store $\mathbf{x}_{in}$ and $\mathbf{y}_{in}$ in an empty row of $\mathcal{B}$.
    }
    {
    Evaluate the loss function at each buffer sample: $\mathcal{L}_i = \ell \left(\mathbf{y}_i, f\left(\mathbf{x}_i\right) \right), \, \forall \left(\mathbf{x}_i, \mathbf{y}_i\right) \in \mathcal{B}$. \\
    Find the buffer element with the minimum loss, at index $i_{\min} = \argmin_i \mathcal{L}_i$. \\
    Replace the element at $i_{\min}$ with $\left(\mathbf{x}_{in}, \mathbf{y}_{in}\right)$.
    }
\textit{\textbf{Model update:}}\\
Evaluate the average loss on the updated buffer $\mathcal{B'}$ as \vspace{-0.8em}\begin{equation*} \mathcal{L} = \frac{1}{N} \sum_{i=1}^{N} \ell \left(\mathbf{y}_i, f\left(\mathbf{x}_i\right)\right). \end{equation*} \\
Perform a backpropagation pass to find \vspace{-0.2em}\begin{equation*} \frac{\partial \mathcal{L}}{\partial w_{j,k}^{(l)}}, \frac{\partial \mathcal{L}}{\partial b_{j}^{(l)}}, \, \forall l \in \{1, \ldots, n_l\}, \forall j, \forall k. \end{equation*} \\
Perform a gradient descent step to update the model parameters. \\
$t \leftarrow t + 1$
}
\end{algorithm}
% \vspace{-1em}

The proposed online training method is motivated by the following observations:
\begin{itemize}
    \item \textbf{Efficient learning}: in its simplest form, stochastic gradient descent performs updates on a set of model parameters $\theta$ as
    \begin{equation}
        \theta_{t} \leftarrow \theta_{t-1} - \eta \nabla_{\theta} \mathcal{L}\left(f, \mathcal{B}_t, \theta_{t-1} \right),
    \end{equation}
    where $\eta > 0$ is a learning rate, $\mathcal{L}$ is a loss function, $f$ is a neural network model associated to $\theta$, and $\mathcal{B}$ is a set of input/output pairs.
    As the gradients are directly proportional to the loss, ensuring that the buffer contains samples with larger errors can lead to faster training. Intuitively, emphasizing the samples with a higher loss allows the training process to focus on the model's largest mispredictions.
    \item \textbf{Data diversity}: a set of streaming measurements is likely to contain highly time-correlated samples, which may result in consecutive model updates on a very limited set of information.
    By retaining samples with a large loss, if the model overfits to a sequence of strongly correlated samples, the buffer will be more likely to retain previous, more diverse samples.
    Additionally, if a rare event occurs, the sample selection mechanism makes it likely that the corresponding samples will be kept in the buffer, as the majority of the buffer will be populated with measurements from a different data distribution.
\end{itemize}

Three state-of-the-art continual learning algorithms have been implemented for a comparative analysis of the proposed method, each of them belonging to a separate family of methods~\cite{DeLange2022}.
The algorithms have been selected due to their widespread use and the possibility of their implementation with a constant memory and computation footprint, allowing for a fair comparison with the proposed method.

\subsubsection{Incremental Classifier and Representation Learning (iCaRL)} memory replay~\cite{Rebuffi2016}.
In the context of classification problems, iCaRL maintains separate data buffers for each target class.
As new data becomes available, these buffers are populated with exemplars, i.e. the samples closest to the means of each class.
In its application to online regression, a single exemplar buffer $\mathcal{P}_t$ is considered, and the loss function becomes
\begin{equation}
    \mathcal{L} = \mathcal{L} \left( \mathcal{B}_t \cup \mathcal{P}_t \right),
\end{equation}
where $\mathcal{B}_t$ is a standard buffer with a first-in-first-out mechanism.
A single sample $p^*$ is added to $\mathcal{P}_t$ from $\mathcal{B}_t$ every $\left| \mathcal{B}_t \right|$ time steps, based on
\begin{equation}
    p^* = \argmin_{\left(\mathbf{x}_i, \mathbf{y}_i \right)\in \mathcal{B}_t} \left|\left| \mathbf{x}_i - \bar{\mathbf{x}} \right|\right|_2,
\end{equation}
where $\bar{\mathbf{x}}$ is the average feature vector over all samples in $\mathcal{B}_t$.
The proposed error-based selection method shares many similarities with iCaRL, as both are based on memory replay mechanisms.

\subsubsection{Online Elastic Weight Consolidation (EWC)} regularization~\cite{Schwarz2018}.
EWC introduces a quadratic term to the loss function, which penalizes parameter changes weighted by the sensitivity of the loss to each parameter, as measured by the diagonal of the Fisher information matrix (in classification).
EWC~\cite{Kirkpatrick2017} requires storing the model parameters $\theta$ and sensitivity factor $F$ for each task, while online EWC~\cite{Schwarz2018} only maintains the $\theta$ and $F$ corresponding to the previous task.
Interpreting the data available at each time step as a separate task, the loss function for online EWC can be expressed as
\begin{equation}
    \mathcal{L} = \mathcal{L}\left( \mathcal{B}_t \right) + \frac{\lambda_{EWC}}{2} \sum_{i=1}^{|\theta|} F_{t-1, i} \cdot \left(\theta_{t, i} - \theta_{t-1, i} \right)^2,
\end{equation}
where $\lambda_{EWC} \geq 0$ is a weighting factor for the penalty term.
Additionally, $F$ is computed as a running sum, given by
\begin{equation}
    F_t = \gamma F_{t-1} + \left(\nabla_{\theta_t}\mathcal{L}_t\right)^2,
\end{equation}
with $F_0$ initialized at zero, and a second hyperparameter $0 \leq \gamma \leq 1$ acting as a discount factor for the penalty updates.

\subsubsection{Learning without Forgetting (LwF)} distillation~\cite{Li2018}.
In a task-incremental setting, LwF stores the model parameters corresponding to each learned task.
During training, the loss function is augmented with a quadratic term penalizing the distance between the predictions with the current set of parameters and the predictions with the stored versions of the model, as obtained with the current set of samples.
In application to online regression, and to maintain a constant footprint, only the model corresponding to the previous time step is employed, resulting in the loss function
\begin{equation}
    \mathcal{L} = \mathcal{L}\left( \mathcal{B}_t \right) +  \frac{\lambda_{LwF}}{N} \sum_{i=1}^N \left|\left| f(\mathbf{x}_{t,i}, \theta_{t}) - f(\mathbf{x}_{t,i}, \theta_{t-1}) \right|\right|_2^2,
\end{equation}
where $\lambda_{LwF} \geq 0$ is a weighting factor.

\subsection{Application to the Case Study}
The online training methods are applied to a fully-connected feedforward neural network model, also known as a multilayer perceptron (MLP).
An MLP contains multiple layers, each applying a linear transformation followed by a nonlinear mapping commonly called \textit{activation function}.
It can be expressed as
\begin{equation}\label{eq:layer}
    \mathbf{a}^{(l)} = g^{(l)} \left(\mathbf{W}^{(l)} \mathbf{a}^{(l-1)} + \mathbf{b}^{(l)}\right),
\end{equation}
where $\mathbf{a}^{(l)}$ is a vector of unit \textit{activations} at layer $l$, $\mathbf{W}^{(l)}$ is the layer's \textit{weight} matrix (indexed as $w^{(l)}_{j,k}$), $\mathbf{b}^{(l)}$ is its \textit{bias} vector (indexed as $b^{(l)}_{j}$), and $g\left(\cdot \right)$ is its activation function.
With this definition, the initial layer activation is equal to the input sample $( \mathbf{a}^{(0)}_i = \mathbf{x}_i )$, and similarly, the output layer activation is equal to the predicted output $( \mathbf{a}^{(n_l)}_i = \mathbf{\hat{y}}_i )$.

The employed activation function for the inner layers is the rectified linear unit (ReLU), defined as
\begin{equation}
    g(x) = \max (0, x) = \begin{cases}
    0 & \text{if } x < 0, \\
    x & \text{otherwise},
    \end{cases}
\end{equation}
and applied element-wise.
As the model is designed for regression, the output layer activation function $g^{(n_l)}\left(\cdot\right)$ is the identity function.
The tunable parameters of the model are the bias matrices and weight vectors of each layer, which are updated via gradient descent.
For this purpose, a commonly used loss function for regression problems is the mean squared error (MSE).
For a single sample, the MSE loss function is defined as
\begin{equation}
    \ell \left( \mathbf{y}, \mathbf{\hat{y}} \right) = \frac{1}{n_o} \sum_{j=1}^{n_o} {\left(y_j - \hat{y}_j \right)}^2,
\end{equation}
where $n_o$ is the number of outputs of the model.
In the presented case study, the regression model is trained to predict a single heat sink temperature value $T_{hs}$.
The loss function is therefore reduced to
\begin{equation}\label{eq:single_sample_mse}
   \ell \left( T_{hs}, \hat{T}_{hs} \right) = {\left(T_{hs} - \hat{T}_{hs} \right)}^2.
\end{equation}
When evaluated on a buffer $\mathcal{B}$ containing $N$ training examples, the MSE becomes
\begin{equation}
    \mathcal{L} = \frac{1}{N} \sum_{i=0}^{N} {\left( T_{hs, i} - \hat{T}_{hs, i} \right)}^2.
\end{equation}
The backpropagation algorithm can then be used to obtain the gradients of the loss function w.r.t.\ each parameter of the model.
For the output layer,
\begin{equation}
    \frac{\partial \mathcal{L}}{\partial {a}^{(l_{out})}} = -2 \left(T_{hs} - {a}^{(l_{out})} \right),
\end{equation}
as $\hat{T}_{hs} = {a}^{(l_{out})}$.
Using the chain rule, the gradients of the loss w.r.t.\ the activations of each hidden layer can be obtained recursively~\cite{Goodfellow-et-al-2016}, as
\begin{equation}
    \nabla_{\mathbf{z}^{(l)}} \mathcal{L} = \mathbf{W}^{(l+1)\top} \nabla_{\mathbf{z}^{(l+1)}} \mathcal{L}\circ g^{(l)}{'} \left( \mathbf{z}^{(l)} \right),
\end{equation}
where $\mathbf{z}^{(l)}$ is the pre-nonlinearity activation at layer $l$, i.e.\ $\mathbf{z}^{(l)} = \mathbf{W}^{(l)} \mathbf{a}^{(l-1)} + \mathbf{b}^{(l)}$ and $g'(\cdot)$ is, in this case, the derivative of the ReLU function.
With the values of $\nabla_{\mathbf{z}^{(l)}} \mathcal{L}$ for every layer, the computation of the gradients w.r.t.\ the weights and biases is simply
\begin{align}
    \nabla_{\mathbf{W}^{(l)}} \mathcal{L} &= \nabla_{\mathbf{z}^{(l)}} \mathcal{L} \cdot \mathbf{a}^{(l-1)\top}, \\
    \nabla_{\mathbf{b}^{(l)}} \mathcal{L} &= \nabla_{\mathbf{z}^{(l)}} \mathcal{L}.
\end{align}
The gradient descent step for the parameters is implemented with stochastic gradient descent (SGD) with momentum, i.e.
\begin{equation}
    \mathbf{W}^{(l)}_t = \mathbf{W}^{(l)}_{t - 1} - \eta \nabla_{\mathbf{W}^{(l)}} \mathcal{L} + \mu \left(\mathbf{W}^{(l)}_{t-1} - \mathbf{W}^{(l)}_{t-2} \right), \, \forall l,
\end{equation}
\begin{equation}
    \mathbf{b}^{(l)}_t = \mathbf{b}^{(l)}_{t - 1} - \eta \nabla_{\mathbf{b}^{(l)}} \mathcal{L} + \mu \left(\mathbf{b}^{(l)}_{t-1} - \mathbf{b}^{(l)}_{t-2} \right), \, \forall l,
\end{equation}
where $\eta$ is the \textit{learning rate} and $\mu$ is the \textit{momentum factor}.
Although there exist alternatives to backpropagation for the training of neural networks, gradient-based models are well-suited to online learning, as the optimizer parameters can be used to control the model parameter dynamics, thus controlling the rate of adaptation of the model to incoming data.

\subsection{Anomaly Threshold Design}

When used to perform anomaly detection, the MLP regression model can be employed to obtain a squared error value corresponding to each sample $i$, as described in (\ref{eq:single_sample_mse}).
In its simplest form, anomaly detection can then be achieved by comparing the value of $\mathcal{L}_i$ to a static threshold value $T$.
With the model trained under healthy operation, prediction errors will exceed the threshold more frequently as degradation occurs.
However, it may be challenging to design a static $T$ ahead of operation, as its value should be dependent on the performance of the trained model.

A more scalable approach, presented here, is to estimate $T$ based on the prediction errors, in an online manner.
Rather than predefining a value for $T$, the designer can decide on a critical value for the probability distribution of the residuals, with the threshold becoming a function of the spread of this distribution.
Under the assumption that the prediction residuals follow a normal distribution centered at zero, their square will follow a $\chi^2$ distribution with one degree of freedom.
Although this assumption does not hold in general, it is approximately valid for unbiased neural networks with a linear function as the output unit activation~\cite{Goodfellow-et-al-2016}.
Using a desired confidence level $\alpha$, the threshold $T$ can then be defined as
\begin{equation}
    T = s^2 \cdot \chi^2_{\alpha}\left(k=1\right),
\end{equation}
where $s^2$ is the unbiased sample variance of the prediction residuals and $\chi^2_{\alpha}\left(k\right)$ is the critical value of the $\chi^2$ with $k$ degrees of freedom at the confidence level $\alpha$.
A recursive algorithm can be employed to obtain $s^2$ without storing additional data.
For example~\cite{Welford1962},
\begin{equation}
    \bar{r}_n = \bar{r}_{n-1} + \frac{r_n - \bar{r}_{n-1}}{n}, \quad \bar{r}_0 = r_0,
\end{equation}
\begin{equation}
    S_n = S_{n-1} + \left(r_n - \bar{r}_{n-1}\right)\left(r_n - \bar{r}_n \right), \quad S_0 = 0,
\end{equation}
\begin{equation}
    s^2_n  =\frac{S_n}{n-1},
\end{equation}
where $r_n = T_{hs, n} - \hat{T}_{hs, n}$ denotes the prediction residual at step $n$ and $\bar{r}_n$ is its mean.
Although the designer must still select a value for $\alpha$, it is not dependent on the residuals, and it has a clearer interpretation than $T$, i.e.\ which proportion of samples should be classified as healthy during the threshold-fitting process.
In this manner, the proposed method can be employed to achieve a self-commissioning anomaly detection system.

% --------------------------------------------------------------------------------
\section{Experimental Results}\label{sec:results}

The experimental verification of the proposed method was carried out on the collected data set, as detailed in Section~\ref{sec:system_description}.
The MLP models are trained to predict heat sink temperature given the last \qty{30}{\min} of rms inverter output current which, when sampled at a resolution of \qty{10}{s}, amount to 180 values per input sample.
The rms current values are filtered on the drive side, reducing the presence of measurement errors.
In the use case presented in this article, where the load under consideration is inductive, the converter output currents are dominated by their low-frequency components.
Therefore, their rms values act as an adequate indicator for temperature prediction.
For the experimental test bench under study, inverter output current proved sufficient to obtain accurate heat sink temperature predictions, since there is a direct relationship between the electrical power of the motor and its stator currents, which neural network models can implicitly learn.
However, more complex applications with additional interacting variables may greatly benefit from their inclusion as model inputs.

The length of the input data window is selected to be long enough to fully represent heating and cooling processes---it is therefore device-dependent and should be tuned according to the monitored variables' time constants.
The number of samples stored in the data buffer ($N$) should be tuned according to the available memory on the computing device.
With 180 inputs and a single output per sample, and a buffer size of 50, the buffer takes up approximately \qty{0.3}{\mega\byte} of memory.
A larger buffer allows for more information to be employed in every model update, and can therefore speed up and increase the stability of the training process.
The training speed and stability of the online training methods are lower than those of the offline methods, as model update steps can only be informed by a portion of the data set, which contains less information than its whole.
In this sense, online training methods under memory constraints can only aim to approximate the performance of a corresponding offline approach, where a complete data set is available during the entirety of the training process, and where batching and shuffling methods ensuring that update steps improve the generalization ability of the model.

The data set is scaled to a $[0, 1]$ range using min-max normalization, with scaling values based on the training set in each iteration of cross-validation.
Other relevant parameters for the online learning models are listed in Table~\ref{tab:parameters}.

\begin{table}[h]
    \vspace{-1em}
        \caption{Parameters of the online training models}\label{tab:parameters}
        \centering
        \begin{tabular}{l c}\toprule
            Parameter & Value \\
            \midrule
            \rowcolor{gray!25}Sampling period & \qty{10}{\s} \\
            Buffer size ($N$) & $50$ \\
            \rowcolor{gray!25}Number of inputs ($n_i$) & $180$ \\
            Number of hidden layers ($n_l$) & $2$ \\
            \rowcolor{gray!25}Hidden layer sizes & $[16, 8]$ \\
            SGD learning rate ($\eta$) & $1\mathrm{e}{-3}$ \\
            \rowcolor{gray!25}SGD momentum factor ($\mu$) & $0.9$ \\
            \bottomrule
        \end{tabular}
    \end{table}

Eight models are trained on the collected data set, with different training methods.
These models are referred to as:
\begin{enumerate}
    \item Offline classifier: trained offline on the complete training data set augmented with data corresponding to anomalous operation. Having been exposed to these additional data, the model expectedly outperforms all regression models in identifying anomalies. However, the collection of anomalous data during field operation is often unfeasible, invalidating this approach. This model is trained with the Adam optimizer using the default parameters suggested in~\cite{Kingma2014}.
    \item Offline regressor: trained offline on the complete training data set, with shuffling and over $100$ epochs. This model is also trained with the Adam optimizer.
    \item Incremental: trained online in a purely incremental manner, i.e.\ with no data buffering.
    \item Buffer: trained using a simple data buffer, where $N$ samples are held in memory and used at every training step. Each new sample overwrites the oldest sample held in the buffer.
    \item Selection: trained using Algorithm~\ref{alg:selection}, where each new sample instead overwrites the sample with the lowest loss.
    \item iCaRL: based on~\cite{Rebuffi2016}. In order to enable a fair comparison with the other methods, each of its two buffers is designed to contain 25 samples.
    \item EWC: based on the online version introduced in~\cite{Schwarz2018}. Its hyperparameters are selected through grid search, with $\lambda_{EWC} = [10, 12.5, 15, 17.5, 20, \mathbf{22.5}, 25, 27.5, 30]$ and $\gamma = [0.7, \mathbf{0.8}, 0.9, 0.95, 0.99]$.
    \item LwF: based on~\cite{Li2018}. Its hyperparameter is selected through grid search with $\lambda_{LwF} = [0.01, 0.05, \mathbf{0.1}, 0.25, 0.5, 0.5, 0.75, 1]$.
\end{enumerate}
The implementation of the models and their training was done on Python 3.10 and with PyTorch 1.12.
The feasibility of the method on edge platforms was also verified on a Raspberry Pi 4B with an ARM Cortex-A72 processor.
On the edge device, the selection method was implemented with TensorFlow Lite 2.11, with a total execution time of \qty{15.7(1.3)}{ms} per inference and training iteration.
This measurement is orders of magnitude below the \qty{10}{\second} base sampling time of the anomaly detection system, which suggests that the method is suitable for devices with far more limited computational capabilities, as well as processes with faster dynamics.

To perform cross-validation, the data set recorded under normal operation is split in 8 parts.
Parts are collected in pairs, with each run using 6 parts for training and 2 parts for testing, resulting in 28 combinations.
During the online training process, the models are continuously evaluated on the test data set, which also includes the recorded anomalous data.

\begin{figure}[t]
    \centering
    \includegraphics[scale=1]{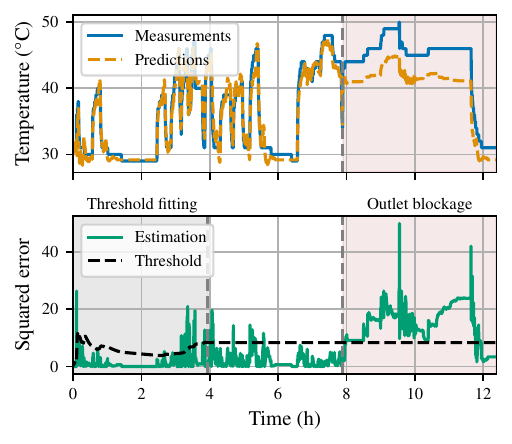}
    \vspace{-0.3cm}
    \caption{Experimental results in a test run, using a model trained with online learning with the proposed sample selection method. In this example, the false positive rate is 3.7\%, and the true positive rate is 85.7\%.}\label{fig:threshold}
\end{figure}

The models are evaluated in terms of their ability to perform anomaly detection, i.e.\ to consistently result in higher prediction errors for anomalous samples.
Samples with a squared error larger than an anomaly threshold parameter $T$ are considered positive predictions, suggesting anomalous behavior.
Fig.~\ref{fig:threshold} shows the heat sink temperature measurements and predictions at the top, with the corresponding squared errors at the bottom, as obtained with a model trained with the proposed buffer sample selection method.
The figure also shows the anomaly threshold, which is fit during the first 4 hours of the experiment as described in Section~\ref{sec:methodology}, and with a confidence level of $\alpha = 99\%$.
During this time, the threshold can be observed to increase with sequences of larger errors, and to decrease with sequences of smaller errors.
The threshold fitting time must be designed to be long enough to accurately capture the variance of the prediction residuals during healthy operation.
After approximately 8 hours, the air outlet is blocked, resulting in larger prediction errors which are in their majority correctly identified as anomalies, as they exceed the now static threshold value.

\begin{figure}[t]
    \centering
    \includegraphics[scale=1]{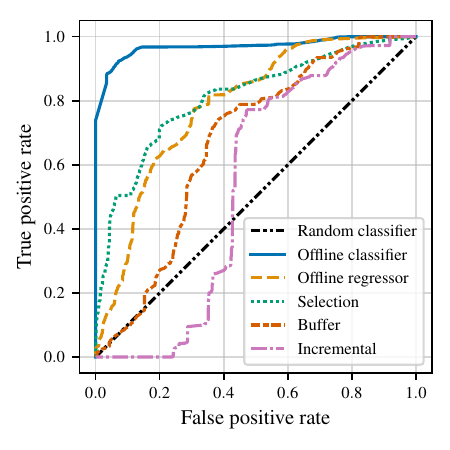}
    \vspace{-0.3cm}
    \caption{Example receiver operating characteristic (ROC) curve for a single training experiment, recorded at an equivalent 11 hours of training.}\label{fig:roc_example}
\end{figure}

\begin{figure*}[t]
    \centering
    \includegraphics[scale=1]{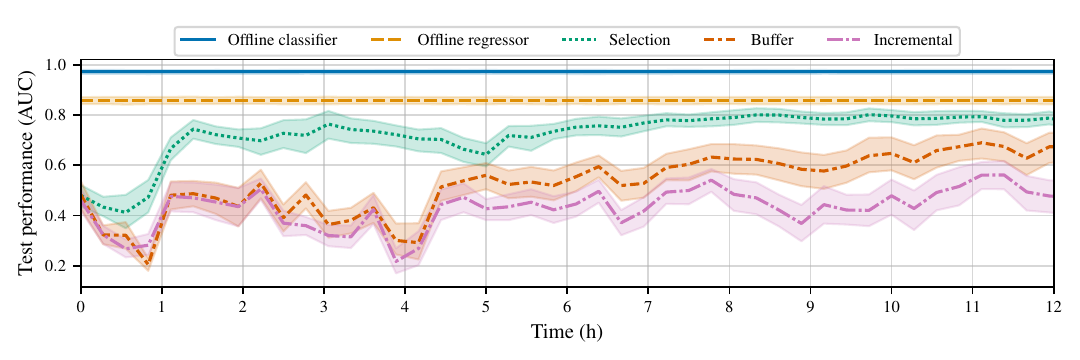}
    \vspace{-0.7cm}
    \caption{AUC results on the test data set, for each of the five considered training methods. The results are graphed as mean values and $95$\% confidence intervals across the 28 cross-validation runs, as a function of training time.}\label{fig:auc_training}
\end{figure*}

\begin{figure}[t]
    \centering
    \includegraphics[scale=1]{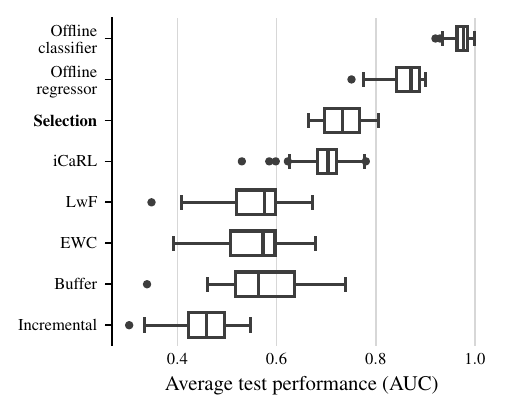}
    \vspace{-0.3cm}
    \caption{Boxplots displaying the average AUC over cross-validation test runs for each considered training method.}\label{fig:auc_mean}
\end{figure}

The classification performance of the models is analyzed by means of receiver operating characteristic (ROC) curves, which represent the true positive rate as a function of the false positive rate, as $T$ ranges from $0$ to $\infty$.
Fig.~\ref{fig:roc_example} illustrates an example set of ROC curves.
These curves are obtained by evaluating the classification performance of the models after approximately 11 hours of training, for each of the five considered methods.
ROC curves can be summarized in a single metric, the area under the ROC curve (AUC), which allows for the comparison of models using a single metric and independently of the selection of a threshold parameter.
AUC can be calculated by numerically integrating an ROC curve.
The maximum value of AUC is 1, corresponding to a perfect classifier, while a random classifier---which predicts the correct class of a given sample with a 50\% probability---achieves an AUC of 0.5.
In the context of this article, the AUC can be understood as a measure of the probability that a randomly selected anomalous sample will result in a higher squared prediction error than a randomly selected healthy sample.
The use of AUC enables a more general analysis than metrics such as accuracy, precision, or recall, since it quantifies the detection performance independently of the value of $T$, which is application-dependent.
The complete results, in terms of AUC as a function of time, are presented in Fig.~\ref{fig:auc_training}.
The results are based on the data collected as explained in Section~\ref{sec:system_description}, i.e.\ under varying torque and speed setpoints and with anomalous data produced by partially blocking the air outlet of the drive's cooling system.
The graphs therefore show aggregated results accounting for the variety of operating conditions under consideration.
These results are further summarized in the boxplots displayed in Fig.~\ref{fig:auc_mean}, which show the distribution of the mean AUC throughout cross-validation iterations, and include additional results corresponding to iCaRL, EWC, and LwF.

The results shown in Fig.~\ref{fig:auc_mean} suggest that online EWC and LwF are not particularly suitable to the case study under consideration, as they are unable to achieve significantly better performance than training with a buffer without a sample selection mechanism.
Nonetheless, the iCaRL method is able to achieve similar results to the proposed error-based sample selection method, with only slightly lower AUC.
As the two methods are closely related, both relying on memory replay with a sample selection mechanism, their similarity should not be surprising.
iCaRL appears to be less consistent throughout cross-validation than the proposed selection method, as suggested by the presence of outliers in its boxplot.
Together with its slightly lower performance, this instability serves as an argument in favor of the proposed method.

\begin{table}[h]
    \vspace{-1em}
    \centering
    \caption{Comparative summary of online training methods}\label{tab:comparison}
    \begin{tabular}{lccc}
    \toprule
    Method      & \begin{tabular}[c]{@{}c@{}}Normalized\\ runtime\end{tabular} & \begin{tabular}[c]{@{}c@{}}Required\\ memory\end{tabular} & Average AUC       \\ \midrule
    \rowcolor{gray!25}Incremental & $1.0$                                              & $|\theta|$                                                & $0.453\pm 0.059$  \\
    Buffer      & $1.14$                                             & $|\theta| + |\mathcal{B}|$                                & $0.570 \pm 0.091$ \\
    \rowcolor{gray!25}Selection   & $1.56$                                             & $|\theta| + |\mathcal{B}|$                                & $0.733 \pm 0.042$ \\
    iCaRL       & $1.18$                                             & $|\theta| + |\mathcal{B}|$                                & $0.692 \pm 0.058$ \\
    \rowcolor{gray!25}EWC         & $3.79$                                             & $3 \times |\theta| + |\mathcal{B}|$                       & $0.560 \pm 0.072$ \\
    LwF         & $2.30$                                             & $2 \times |\theta| + |\mathcal{B}|$                       & $0.555 \pm 0.076$ \\ \bottomrule
    \end{tabular}
\end{table}

Table~\ref{tab:comparison} summarizes the computational requirements and the average performance of each online training algorithm.
Each runtime is measured on a Raspberry Pi 4B and reported as normalized with respect to the incremental case.
Memory requirements are expressed as the ideal minimum number of parameters required for implementation.
The iCaRL method is remarkably capable of significantly improving the AUC over the simple buffer method, without additional memory requirements and with almost no increase in runtime.
The proposed selection method requires a somewhat longer runtime, as it requires an additional forward evaluation of the model, but outperforms every other considered method in terms of AUC.

The results show that, out of the six considered online training methods, the proposed method is able to achieve the closest results to the two offline models, with higher values of AUC, a faster training process, and higher stability across cross-validation iterations.
The model trained purely incrementally, without buffering, does not obtain a better performance than a random classifier, which theoretically achieves an AUC of $0.5$.
Although the training with a simple buffer often achieves higher values of AUC, they are still significantly lower than those achieved with the proposed method, and importantly, are substantially less stable, as shown by the wider bands in the corresponding plot of Fig.~\ref{fig:auc_mean}.
The iCaRL method, as another algorithm based on memory replay with a sample selection mechanism, is able to achieve the closest performance to the proposed selection method and presents a minimal computational footprint.
However, it may require more careful consideration in its design than the proposed method, as the buffer of exemplars must be defined and maintained in conjunction with the rolling window buffer.

% --------------------------------------------------------------------------------
\section{Conclusion}
This article presents a new method for the online training of neural network models that improves on standard buffering without introducing additional memory requirements.
The proposed method is applied to the detection of thermal anomalies in a frequency converter, as part of a motor drive test bench, and achieving higher classification accuracy than the considered online alternatives.

The performance of the different training methods is compared in terms of area under the ROC curve (AUC), where samples are categorized individually, to avoid a loss of generality.
However, the prediction errors are time-correlated, and their relationship may be exploited to further improve anomaly detection accuracy.
Final decisions on whether a series of errors classifies as anomalous can be achieved by means of normality tests or employing more advanced techniques, such as the cumulative sum control chart (CUSUM) method.
In any case, a threshold or cut-off parameter would have to be defined for anomaly detection in a production environment.

In online learning contexts, training machine learning models with gradient descent offers a clear advantage over direct fitting methods, as learning rates and other hyperparameters (e.g.\ number of epochs, mini-batch size) allow for the tuning of the parameter dynamics of the model.
Through this tuning process, designers can balance the trade-off between a faster response, i.e.\ quicker adaptation to the current data buffer, and a slower response, i.e.\ better extrapolation to data in previously stored buffers.

The modeling of time-series data with incremental machine learning introduces additional trade-offs in the design of the data buffer, which were not directly addressed in the present study.
Consecutive samples can be stored in less memory than non-consecutive ones; thus, there exists a trade-off between storing samples consecutively and ensuring data diversity.
A data buffer containing consecutive data points will result in an increased probability of training with similar samples---and thus increase the model's proneness to overfitting.

Achieving powerful, low-footprint, and reliable methods for online machine learning may reduce the current reliance on cloud platform services and large-scale data storage, thus enabling the use of data-driven modeling in a wide variety of applications.
Self-commissioning condition monitoring algorithms are a clear example of one such application, as machine learning models can benefit from online training to adapt to individual operating conditions, while simultaneously requiring minimal knowledge of the system and its environment.
Other possible uses of online machine learning in the area of power converters and drives include adaptive control algorithms, efficiency optimization, and load forecasting, among others.
These methods may be applied to variable frequency drives in systems such as motors, pumps, fans, and compressors, as well as other power converter systems.

Besides application-focused development, future research directions may explore the integration of additional improvements to online training---i.e.\ regularization, ensemble methods, flexible architectures---within resource-constrained environments.

% References
\bibliographystyle{Bibliography/IEEEtranTIE}
\bibliography{bibliography} %IEEEabrv instead of IEEEfull

\vspace{-2.5em}
\begin{IEEEbiography}[{\includegraphics[width=1in,height=1.25in,clip,keepaspectratio]{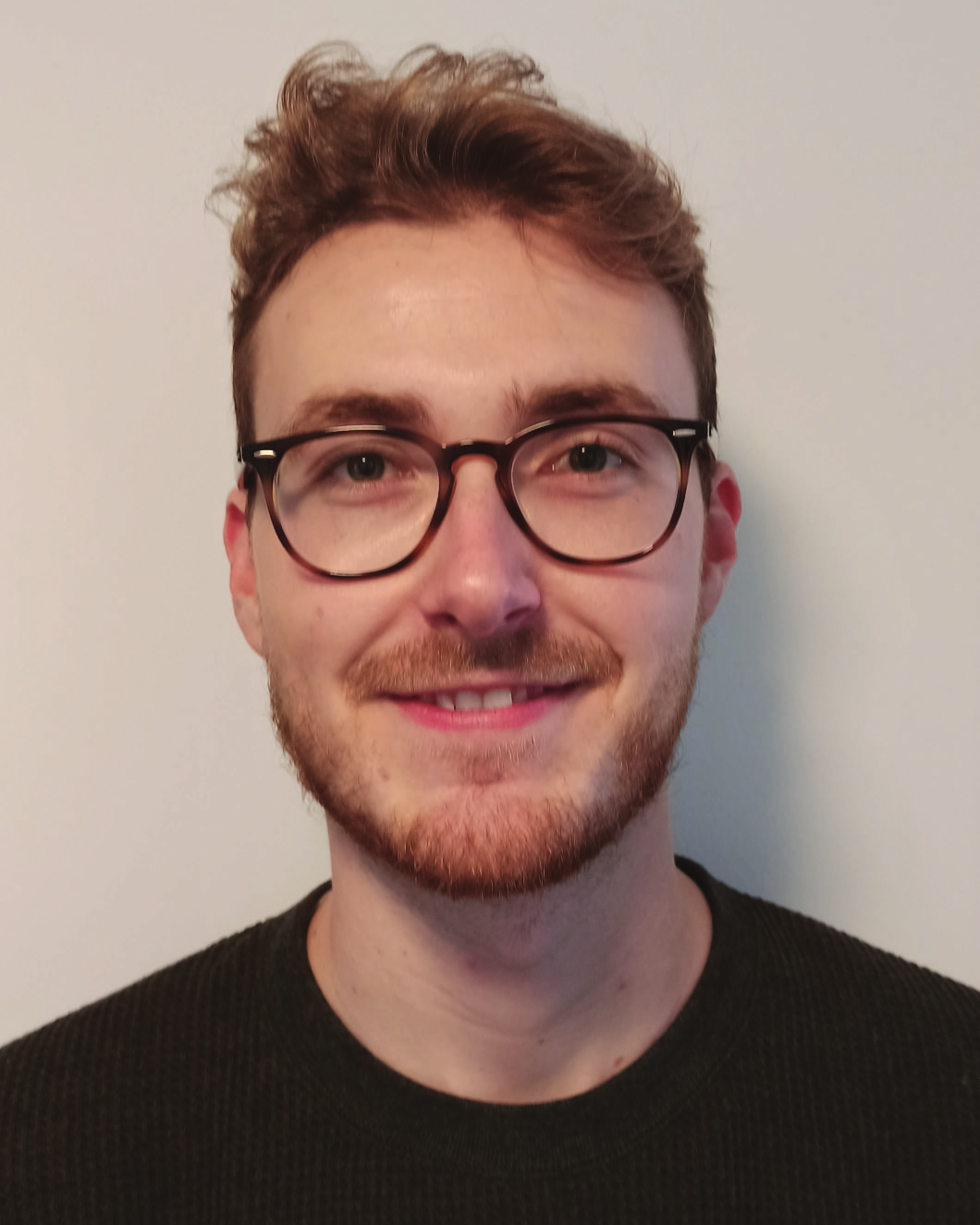}}]
{Pere Izquierdo G\'omez} (S'22) received the B.Sc. degree in energy engineering from the Polytechnic University of Catalonia, Barcelona, Spain, in 2018, and the M.Sc. degree in energy engineering from Aalborg University, Aalborg, Denmark, in 2020. He is currently pursuing the Ph.D. degree in the Technical University of Denmark (DTU), as part of the Department of Wind and Energy Systems, working on power converter diagnostics and control with a focus on machine learning applications.
\end{IEEEbiography}

\vspace{-2.5em}
\begin{IEEEbiography}[{\includegraphics[width=1in,height=1.25in,clip,keepaspectratio]{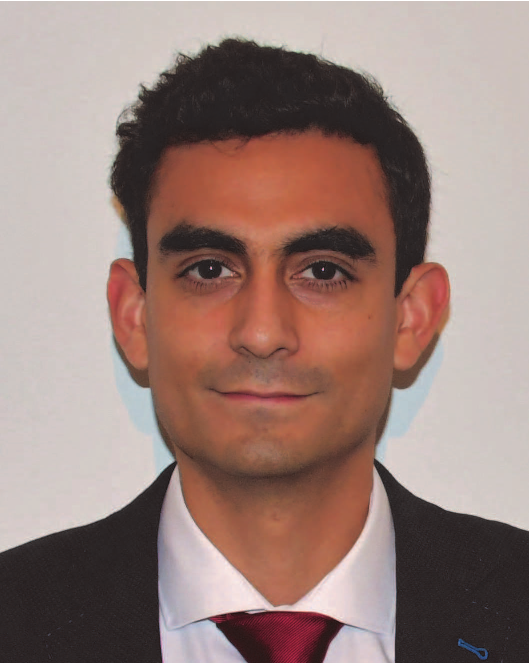}}]
{Miguel E. L\'opez Gajardo} earned the Ingeniero Civil Electr\'onico degree and M.Sc. in Electronic Engineering from the Universidad T\'ecnica Federico Santa Mar\'ia, Valpara\'iso, Chile, in 2014. He worked in the Advanced Center for Electrical and Electronic Engineering (AC3E) as R\&D Engineer from 2015 to 2020. He is pursuing a Ph.D. degree at the Technical University of Denmark (DTU), as part of the Department of Wind and Energy Systems.
His research interests include power electronics, e-mobility, renewable energy conversion systems, drives, AI, and their applications.
\end{IEEEbiography}

\vspace{-2.5em}
\begin{IEEEbiography}[{\includegraphics[width=1in,height=1.25in,clip,keepaspectratio]{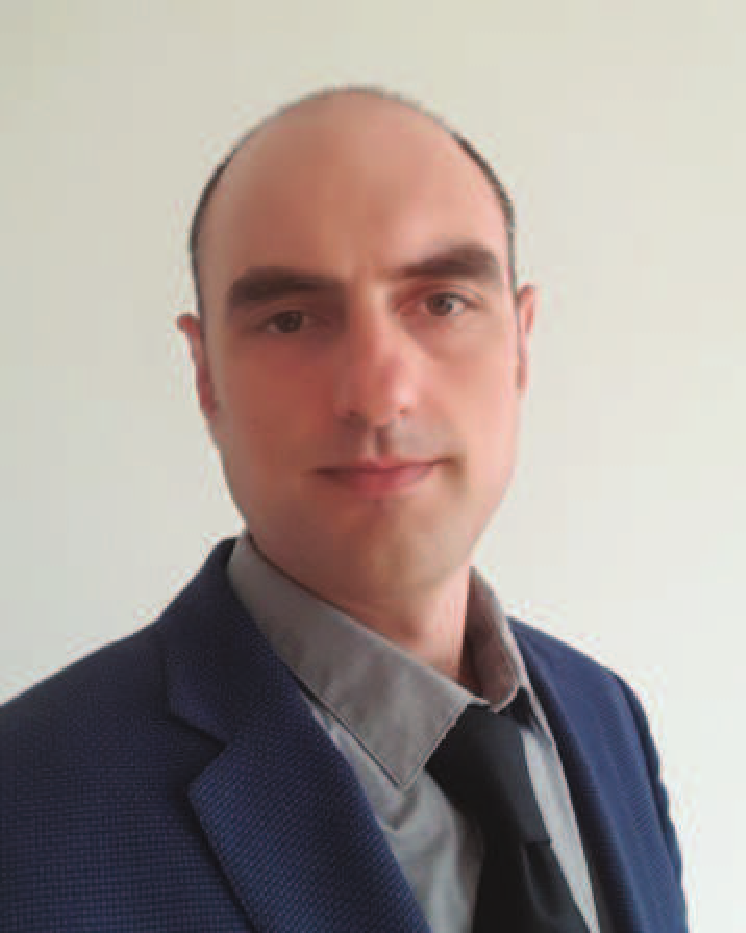}}]
{Nenad Mijatovic} after obtaining his Dipl.Ing. education in Electrical Power Engineering at University of Belgrade, Serbia in 2007, was enrolled as a doctoral candidate at Technical University of Denmark. He received his Ph.D. degree from Technical University of Denmark for his work on technical feasibility of novel machines and drives for wind industry. Upon completion of his PhD, he continued work within the field of wind turbine direct-drive concepts as an Industrial PostDoc. Dr. N. Mijatovic currently holds position of Associate Professor at Technical University of Denmark where he is in charge of managing research projects and education related to the field of electrical machines and drives, power electronic convertors,  motion control, application of energy storage and general applications of low frequency electromagnetism and large scale application of superconductivity with main focus on emerging eMobility and renewable energy generation. He is a member of IEEE since 2008 and senior member of IEEE since 2018 and his field of interest and research includes novel electrical machine drives/actuator designs, operation, control and diagnostic of electromagnetic assemblies, advance control of drives and grid connected power electronics, energy storage and eMobility.
\end{IEEEbiography}

\vspace{-2.5em}
\begin{IEEEbiography}[{\includegraphics[width=1in,height=1.25in,clip,keepaspectratio]{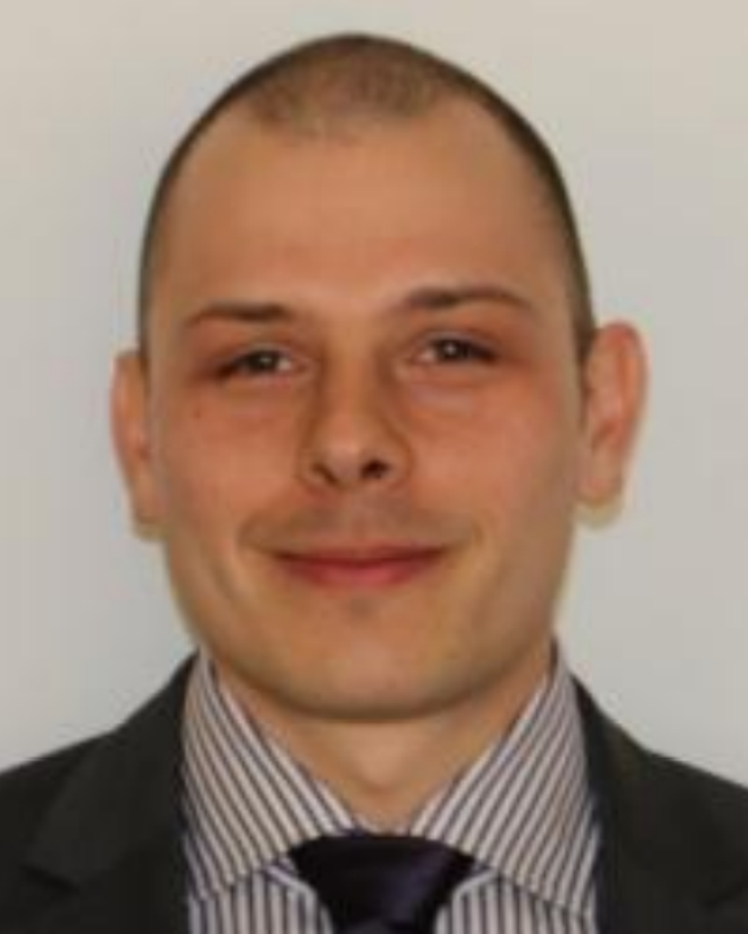}}]
{Tomislav Dragi\v{c}evi\'{c}} earned his M.Sc. and industrial Ph.D. degrees in electrical engineering from the Faculty of Electrical Engineering, Zagreb, Croatia, in 2009 and 2013, respectively. He is currently a professor at Technical University of Denmark, where he leads research in the digitization of power converters. He has published more than 150 IEEE journal articles in these areas. He serves as associate editor of IEEE Transactions on Industrial Electronics and IEEE Emerging and Selected Topics in Power Electronics and an editor of IEEE Industrial Electronics Magazine. He is a Senior Member of the IEEE.
\end{IEEEbiography}

\end{document}